# Development of Application-Specific Large Language Models to Facilitate Research Ethics Review

Sebastian Porsdam Mann,[1,2,3] Joel Jiehao Seah,[3]
Stephen R. Latham,[4] Julian Savulescu,[3,6*] Mateo Aboy,[5] Brian D. Earp[3,6]

1. Centre for Advanced Studies in Bioscience Innovation Law (CeBIL), Faculty of Law, University of Copenhagen, Karen Blixens Pl. 16, 2300 Copenhagen, Denmark
2. Faculty of Law, University of Oxford, St Cross Building St. Cross Rd, Oxford OX1 3UL, United Kingdom.
3. Centre for Biomedical Ethics, Yong Loo Lin School of Medicine, National University of Singapore, 10 Medical Dr, #02-03 MD 11, Singapore 117597.
4. Yale Interdisciplinary Center for Bioethics, Yale University, 238 Prospect St, New Haven, CT 06511, USA.
5. Centre for Law, Medicine, and Life Sciences (LML) & Centre for Intellectual Property and Information Law (CIPIL), Faculty of Law, University of Cambridge, 5 West Rd, Cambridge CB3 9DP, United Kingdom
6. Uehiro Centre for Practical Ethics, Faculty of Philosophy, University of Oxford, 16-17 Saint Ebbe's St, Oxford OX1 1PT, United Kingdom.

*Corresponding author. E-mail: jsavules@nus.edu.sg Telephone: +65 6601 1207

## Abstract

Institutional review boards (IRBs) play a crucial role in ensuring the ethical conduct of human subjects research, but face challenges including inconsistency, delays, and inefficiencies. We propose the development and implementation of application-specific large language models (LLMs) to facilitate IRB review processes. These IRB-specific LLMs would be fine-tuned on IRB-specific literature and institutional datasets, and equipped with retrieval capabilities to access up-to-date, context-relevant information. We outline potential applications, including pre-review screening, preliminary analysis, consistency checking, and decision support. While addressing concerns about accuracy, context sensitivity, and human oversight, we acknowledge remaining challenges such as over-reliance on AI and the need for transparency. By enhancing the efficiency and quality of ethical review while maintaining human judgment in critical decisions, IRB-specific LLMs offer a promising tool to improve research oversight. We call for pilot studies to evaluate the feasibility and impact of this approach.

Research involving human participants has the potential to yield invaluable knowledge that can advance medical practice, improve health outcomes, and inform us about societal practices and human behaviour. However, history has shown that such research also carries risks for participants who may be subjected to exploitation or unjustifiable harm if adequate protections are not in place.[1,2] Consequently, a robust framework of ethical norms, codes of conduct, regulations, and laws has emerged to guide and govern human subjects research.[3-7]

A key component of the ethical oversight infrastructure (i.e., Human Research Protection Programs) is the institutional review board (IRB), also known as a research ethics committee. Before research involving human participants can proceed, the study protocol must undergo review and approval (and potentially also monitoring post-approval) by an IRB to ensure regulatory compliance, sound scientific design, and ethical acceptability. In theory, the IRB review process is an essential safeguard to protect research participants from excessive risk and unjustifiable harm. In practice, however, it suffers from several well-documented problems and limitations.[8] The net result of these problems is an ethics review process often marked by delays, unpredictability, and insufficient accountability - issues which can impede valuable research without necessarily improving protections for human participants.

For example, studies have found wide variability in how different IRBs interpret regulations, make value judgments, and reason through ethical issues.[9,10] There are also significant discrepancies in the levels of scrutiny applied, with some protocols (based on the IRB's determination of the study's level of 'minimal risk' and its risk appetite) undergoing 'Full Board' review while very similar studies are classified as 'Exempt' or eligible for 'Expedited' review. This inconsistent application of standards is compounded by an apparent lack of transparency in many cases (i.e., IRB deliberations and decisions appear to be shrouded in secrecy due to the lack of adequate rationale in its communications to researchers, or from the absence of sufficient and accurate documentation of the IRB's reviews).[11] And there is often no robust mechanism for an institutional memory of past IRB discussions and decisions.

In light of these limitations, it has recently been proposed to make use of advances in generative artificial intelligence (AI), in particular large language models (LLMs), as a *first pass* or *adjunct* screening tool to facilitate the speed with which IRBs can assess research proposals.[12,13] While general LLMs such as those employed in OpenAI's GPT, Meta's Llama or Anthropic's Claude series have shown initial promise and potential in this area,[12,13] their use as screening tools is also attended by risks and potential ethical concerns. Ethical review is a paradigmatic case of an area in which *human* oversight and judgment are widely believed to be of paramount importance.[14] This belief is likely strengthened by an awareness of known issues with LLMs related to accuracy, bias, explainability, consistency, reproducibility, and the potential for LLMs to 'hallucinate' (i.e., the ability of LLMs to generate plausible-sounding outputs that are not directly grounded in factual or training data due to the generative AI models' capacity to create, synthesize, or extrapolate information probabilistically based on its learned patterns).[15,16] Moreover, most IRBs, unlike most off-the-shelf LLMs, are institution-specific: there may be particular institutional requirements that need to be followed based on local factors, including local regulations, or specific concerns of the local population, which are germane to ethical review decisions.

For these reasons, any proposal to cede or delegate important aspects of ethical oversight to AI is understandably controversial. We readily acknowledge these concerns. One way to address them, as has previously been proposed, is to use these systems not as a replacement but rather as a supplement to human review.[12] In this paper, we build on this suggestion by proposing the augmentation of human review with application-specific IRB LLMs designed and validated specifically for this purpose. This design can be achieved through tried-and-tested technical means such as application-specific fine-tuning, Retrieval-Augmented Generation (RAG), and prompt-engineering.[17] To the extent that the design, implementation, and rigorous validation of an application-specific IRB AI is successful, the resulting LLM will have access to additional and higher weighted knowledge specific to individual IRBs which partially addresses concerns related both to accuracy and to the need for context sensitivity. Such purpose-built LLMs could handle pre-submission checks and initial reviews, freeing up time and attention for more complex, sensitive, or high-risk cases; however, difficult judgment calls and final determinations would still rest with human decision-makers.

In the sections below, we develop this argument in greater detail. We begin by providing background information on the need for, as well as known problems with, IRB review, as well as briefly detailing existing proposals for the use of LLMs to address these problems. Next, we provide a concise technical overview of LLMs generally and of application-specific IRB LLMs specifically, illustrating their potential with examples from the burgeoning literature of personalized LLMs for bioethical applications. Building on these technical foundations, we go on to present an operational vision for IRB-specific LLMs and walk through how they could be integrated into existing workflows to improve speed, consistency, and quality. Finally, we consider some of the key limitations and open questions related to our proposal, charting an agenda for future research and implementation efforts.

**II. Problems with ethical review of human subjects research**

The primary function of IRBs is to ensure that research protocols comply with applicable laws and regulations, have sound scientific design, and are ethically acceptable. This involves a careful assessment of the risks and anticipated benefits of the proposed research, as well as a determination of whether the rights and welfare of participants are adequately protected. IRBs also have a responsibility to ensure that consent will be obtained in a manner that is free from coercion or undue influence, and that participants are presented with adequate information about the study.

While the requirement of IRB review has certainly played a significant role in the furtherance of research ethics, the process can subject proposed research to significant, albeit unnecessary delays which may in turn slow down scientific progress and hinder advances in medical treatment.[18]

These delays stem from multiple sources. Many IRB offices face significant challenges in institutional funding and staffing, resulting in inadequate human resources and overworked offices with extended turnaround times.[19] In our own experience, poor quality submissions exacerbate delays, often lacking key details, containing discrepancies, missing documents, or using excessive technical jargon in participant-facing materials. Delayed researcher responses to IRB queries further prolong the process. Unclear and complex IRB communications, especially regarding required modifications and their rationales, can lead to misunderstandings. Research in jurisdictions with complicated legal landscapes, such as the United States with its federal and state laws, or in international multi-site studies adds complexity. Critically, the lack of institutional memory - specifically, the absence of a searchable database of past IRB decisions and outcomes - hinders efficient reviews and contributes to variability in IRB interpretations and decisions.[20,21] Additionally, inadequate procedures for the secure handling of participants' data often complicate the approval process, further extending the time required for IRB review and approval.

In light of these challenges, there is a clear need for innovative solutions that can enhance the speed, consistency, and quality of the IRB review process. One promising approach is to leverage the power of generative AI, particularly LLMs, to assist with the initial screening and evaluation of research protocols.

**III. Generative AI for IRB review**

LLMs are general-purpose AI models whose primary function is to predict the next token (a short word or a fragment of a word) in a sequence of natural language. Recent advances in data availability, hardware, and model architecture have greatly expanded the capacities of multi-modal generative AIs to process text and other modalities, such as sound, image, and video.[22] At the time of writing, state-of-the-art LLMs are trained on huge corpora of text and show significant performance across a wide variety of information tasks.[23] Recently, the introduction of reasoning models or 'reasoners' (e.g., OpenAI's o1 and o3 series, DeepSeek-r1) now confer LLMs the capability to spend more time processing a request, breaking it down into a series of steps before providing a response.[1] This 'Chain-of-Thought' (CoT) technique – an automated process which directs the AI to approach the problem/user's prompt in a stepwise fashion – can be used to generate outputs that purportedly surpass traditional LLMs (e.g., GPT-4o) significantly, in tasks that are reasoning-heavy. In essence, when the LLM is directed to 'think' – and when it 'thinks' longer – the better its responses would generally be. The development of reasoners has also led to advancements in agentic AIs – AIs which upon given a goal, have the ability to achieve that goal – that is, break it down into necessary steps and executing these – with no further human input.[2]

---

[1] See generally: https://openai.com/index/learning-to-reason-with-llms/
[2] Novel real-world applications of AI agents include OpenAI's Operator and Deep Research. Operator is an agent that browses the internet to autonomously execute human tasks (e.g., filling up forms, ordering groceries), while Deep Research is an agent that performs 'multi-step' research by aggregating and 'synthesizing' content from various

LLMs are increasingly used in biomedical contexts, including for interpreting and generating clinical notes and patient reports, to assist in limited areas of diagnostics and treatment, as well as in various administrative and research-oriented tasks.[24] Consequently, the idea of using LLMs to screen protocols submitted to IRBs has been raised.[12] Indeed, an early study demonstrated the potential of this idea across a series of case studies using four different LLMs, each of which was able to pick up on several ethical issues associated with each of seven case studies.[12] However, the study also noted several key omissions, concluding that human oversight continues to be important and that LLMs should be used only as adjunct screening tools.[12]

While the use of LLMs as screening tools prior to a human review is promising, it is also associated with several practical and ethical concerns. Off-the-shelf LLMs trained on generic datasets sometimes lack the domain-specific knowledge and contextual understanding necessary to make nuanced judgments about the ethical acceptability of research protocols. They may also be prone to biases and blind spots that could lead to unfair or unreliable assessments, make mistakes, and miss important areas of potential ethical concern.[12] Moreover, the "black box" nature of traditional LLMs' internal mechanisms makes it difficult to interrogate their reasoning *a priori* and ensure that their potential outputs will align with relevant ethical principles and regulatory requirements. However, with the advent of reasoners, we recognise that newer LLMs can now provide interpretability of its responses (which in the process, also ameliorate concerns regarding hallucinations). By making its CoT available to users, the LLM's thought processes can be interrogated through human evaluation and validation of the logic, veracity, depth, and breadth of reasonings.

In addition to these practical issues, many people may experience general discomfort with the notion of ethical review carried out – perhaps to any extent – by AI systems. Ethical deliberation is a complex process involving not only pure analysis and computation but also qualities such as situational awareness, empathy, and nuanced balancing of competing ethical concerns. There is a legitimate worry that relying too heavily on AI systems, even as a screening or support tool, could lead to a diminishment of these essential human qualities in the review process. IRB members may feel less inclined to engage deeply with the ethical dimensions of research proposals if they believe that an AI system has already done the heavy lifting for them. Over time, this could lead to a gradual erosion of the ethical skills and sensitivities that are so crucial to the integrity of the review process.

Unlike other areas where AI is being deployed, such as medical diagnosis or financial risk assessment, the judgments made by IRBs are not purely empirical or technical in nature. They involve weighing competing moral considerations, interpreting abstract principles in light of concrete circumstances, and making difficult trade-offs between individual rights and interests and potential societal benefits. These are quintessentially human tasks that require a deep understanding of the ethical frameworks and values that undergird the research enterprise. Delegating such tasks to AI systems, even partially, may be seen as an abdication of moral responsibility on the part of IRBs.[25]

Another critical concern is the potential inability of LLMs to fully account for the local institutional and cultural contexts in which IRBs operate. While the basic principles of research ethics, such as respect for persons, beneficence, and justice, are widely applicable, their interpretation in practice can vary significantly depending on the specific setting and population involved. For example, what constitutes an acceptable level of risk or an appropriate consent process may differ depending on the cultural norms, educational backgrounds, and socioeconomic conditions of research participants. Similarly, the institutional mission, resources, risk tolerance, and policies and procedures of a particular IRB may shape its approach to protocol review in ways that are not easily captured by a generic AI model.

This context-specificity is one of the reasons why IRBs are typically localized within particular institutions or research communities, rather than being centralized at a national or international level. Especially in the case of IRBs overseeing social and behavioural research, they are expected to have a deep understanding of the local research landscape and to be attuned to the needs and concerns of the populations they serve.

---

online sources. Leveraging OpenAI's o3 reasoning model, it interprets, 'reasons', and analyses the gathered information to generate, within minutes, a comprehensive fully cited report with reportedly high-quality references (e.g., an up-to-date detailed report on gene therapies that have secured regulatory approval for condition X).

Thus, the current situation may be summarized as follows. IRB review serves an important function yet the conditions necessary for this function to be effectively fulfilled are often not available to those charged with carrying out IRB review. As a result, the quality of reviews may often be lacking, while much beneficial research is delayed and otherwise hindered. The use of LLMs to assist in pre-review screening (e.g., for the completeness of the IRB application, informational consistency across the document, and alignment with ethical norms and legal requirements prior to ethical review) has been proposed as a means of addressing this delay. While promising, such use is attended by significant concerns partially related to accuracy and partially related to more general worries about the nature of ethical reasoning and the need for review that is sensitive to the local context.

In the sections that follow, we suggest that the use of application-specific LLMs trained on or given access to previous local IRB queries (i.e., questions raised by the IRB to researchers), as well as IRB committee discussions and decisions, institutional and IRB-specific policies and procedures, and local regulations, could both increase accuracy and significantly ameliorate the bulk of these concerns.

**IV. Application-Specific IRB LLMs**

LLMs achieve their general-purpose capabilities due to the vast and diverse datasets used in their initial training. However, when high performance is desired within a specific domain rather than for general tasks, the effectiveness of LLMs can be significantly enhanced through a process known as fine-tuning. Fine-tuning involves further training an already pre-trained LLM on a targeted, specialized dataset relevant to the specific area of interest. This process refines the model's understanding by adjusting its internal weights to align more closely with the patterns, terminology, reasoning, knowledge base, and nuances of the specialized data. As a result, fine-tuned AI models become more adept at generating accurate and contextually relevant outputs within their application-specific domains, typically surpassing the performance of general-purpose models in tasks such as medical diagnosis, legal analysis, and other field-specific applications.[26,27]

Application-specific fine-tuning helps reduce hallucinations in LLMs by aligning the model's outputs more closely with domain-specific knowledge. This can be further enhanced through RAG: a strategy that grounds the model's outputs in external, reliable data sources (i.e., domain-specific reference information).[28] RAG works by using embeddings, which are numerical representations that encode text into structured vectorized forms, allowing the model to perform efficient computational searches across vast datasets. This structured approach enables the model to locate and retrieve specific, contextually relevant information during text generation, acting as an external memory or knowledge base. By grounding the model's responses in actual data, RAG helps ensure that outputs are more accurate and reliable. Practical examples of RAG include company AIs that access internal documents to provide precise answers to user queries, thereby reducing the risk of generating incorrect or fabricated information.[29]

We use the term 'application-specific LLMs' to refer to LLMs which have been 1) pre-trained, 2) fine-tuned for a specific application domain (i.e., IRB reviews), 3) enhanced with domain-specific information using RAG, and 4) customized for a specific application through prompt-engineering. The use of application-specific LLMs has the potential to address several of the specific concerns raised above with respect to the use of general LLMs in IRB review. Recently, a growing body of literature has investigated the use of a specific type of LLMs within bioethics, including LLMs which use for their training data or knowledge-base information produced by, or characterizing, a specific individual.[30] Examples of such *personalized* LLMs include models fine-tuned on an individual's previously published works, enabling them to produce text more in line with the style, format, and argumentation used by that individual than a base model[31,32] and similar models fine-tuned on relevant data for individual patients which might be used to predict that patient's preferences in cases of medical incapacity.[33]

Building on these preliminary examples, we envision the development of IRB-specific LLMs that are designed and validated to the unique needs and institutional contexts of individual review boards. These models would be fine-tuned on IRB-specific datasets. They would also be equipped with retrieval capabilities that allow them to access and incorporate the latest relevant information from knowledge-bases established for that purpose. Note that while we focus here on individual IRBs, our proposal of using application-specific rather than generic LLMs for IRB review is also germane to centralised or supra-institutional review boards.[34]

The process of developing an IRB-specific LLM would involve several key steps:

1. **Data collection and curation**: The first step would be to gather a comprehensive dataset of materials relevant to the IRB's work, including IRB literature (e.g., fundamentals about clinical trial design and good practices, IRB and bioethics peer-reviewed literature), past protocol submissions and related documents (e.g., consent forms, study advertisements), decision letters and IRB queries (i.e., IRB submission or review-related communications), meeting minutes, institutional and IRB policies and procedures, and relevant laws and regulations. This dataset would need to be carefully curated to ensure its quality, with attention to issues such as bias and confidentiality. Besides key IRB documentation, recommendations and guidelines from national ethics bodies and/or expert advisory committees (e.g., national bioethics committees, ethics think tanks) could also be included in the model's database for retrieval-augmentation.
2. **Model selection:** The next step involves selecting an appropriate existing general LLM to serve as the base model for fine-tuning and deployment. Available options include proprietary models such as Anthropic's Claude and OpenAI's GPT and o-series, as well as open-source alternatives like Meta's LLaMA, or DeepSeek's v3 and R1 models. The choice of model should be guided by several key considerations, including performance, scalability, and compatibility with the institution's technical infrastructure and data security requirements. Smaller, open-source models have the advantage of being deployable on local servers, offering enhanced control over data handling and partially mitigating confidentiality concerns. This local deployment option can be particularly beneficial when sensitive IRB-related data is involved, as it allows institutions to maintain tighter control over data flow and access. The final model selection should align with the IRB's institutional policies, including those outlined in its AI governance framework or policy, to ensure compliance with requirements for data processing, security, and ethical standards. Additionally, the selected model must support the IRB's specific needs, such as the ability for enterprise-quality data protection, application-specific fine-tuning, and the ability to integrate with RAG to access and utilize domain-specific knowledge efficiently. Ultimately, the chosen model should not only meet technical criteria but also adhere to the IRB's broader institutional and ethical mandates, ensuring responsible use of AI within the context of research ethics review.
3. **IRB application-specific fine-tuning:** The fine-tuning process can be divided into two key stages: (1) creating a general application-specific IRB LLM, and (2) further training this IRB LLM to align with the specific institutional context. In the first stage, the base model is fine-tuned on a comprehensive IRB-specific dataset, incorporating general legal standards, regulations, best practices, and relevant peer-reviewed literature. This step equips the model with a deep understanding of the broader IRB landscape, enabling it to recognize, 'reason', and generate outputs that reflect the unique language, terminology, and ethical frameworks commonly used in IRB decision-making. In the second stage, the model undergoes additional fine-tuning to adapt specifically to the institution's context. This involves training the IRB LLM on high-quality text describing the institution's specific IRB processes, policies, historical decisions, and other relevant documentation. The goal is to imbue the model with the distinct language patterns and contextual nuances that capture the institution's decision-making criteria, ethical considerations, and procedural standards. This institution-specific training step ensures that the model's CoT/reasoning and outputs are not only aligned with general IRB practices but also tailored to reflect the unique values and operational specifics of the institution. Together, these steps enable the fine-tuned IRB LLM to produce highly relevant, context-aware outputs that closely align with both general and institution-specific decision-making frameworks, enhancing its effectiveness as a support tool in the ethical review process.
4. **Retrieval-augmentation (RAG)**: To enhance the model's ability to provide up-to-date and contextually relevant information, it would be equipped with retrieval capabilities using embeddings. This would involve creating structured vector representations of key documents in the IRB's factual knowledge base, such as past decisions, policies, and guidelines. During the generation process, the model could then efficiently search and retrieve relevant information to inform its analyses and recommendations.
5. **Testing and validation:** Before being deployed in a live IRB setting, the application-specific LLM would need to undergo rigorous testing and validation to ensure its accuracy, reliability, fairness, and adherence to ethical and regulatory principles. This would involve both quantitative evaluations of the model's performance on benchmark tasks, as well as qualitative assessments by IRB staff, members, and other stakeholders such as legal counsel to evaluate the model's practical utility, fairness, and alignment with ethical principles. These evaluations would focus on the model's ability to: provide logical and high-quality reasoning with sufficient depth and breadth; generate accurate, contextually relevant outputs; identify potential biases; and ensure compliance with applicable laws and regulations. Feedback from these assessments should guide iterative refinements, ensuring that the model meets the rigorous standards required for responsible use in IRB decision-making and review processes. By comparing the model's reasonings and outputs to previous human IRB decisions on similar cases, evaluators can assess the accuracy, consistency, and alignment of the AI's recommendations with established IRB standards and practices.
6. **Post-deployment audits and continuous-improvement through reinforcement learning from human feedback**: After deployment, the LLM would be subject to periodic audits to monitor its performance and ensure ongoing alignment with IRB standards. This process would involve systematically comparing the LLM's reasonings and recommendations with the IRB's deliberations and final decisions. Differences between the initial LLM output and the final IRB decision, which integrates human judgment in addition to the IRB-AI, can serve as valuable data for continuous improvement. This before-and-after analysis allows for high-quality reinforcement learning through expert human feedback, enabling the model to learn from expert corrections and adjustments made during IRB reviews. Such feedback loops are essential for refining the application-specific LLM, addressing deficiencies or biases in the model's training data, and enhancing its decision-making capabilities over time. The audit process would also incorporate direct user feedback, focusing on scenarios where human reviewers diverged from the LLM's recommendations, the frequency of these instances, and the reasons behind them. This structured feedback mechanism not only identifies the model's limitations but also guides iterative retraining, ensuring the LLM evolves to better support IRB decision-making while maintaining ethical integrity and regulatory compliance

Once developed and rigorously validated, the IRB-specific LLM could be integrated into the review workflow in various ways to enhance the efficiency and quality of the process. For example:

a. **Pre-review screening:** The LLM (or AI agents similar to 'Operator') could be used to autonomously screen incoming protocol submissions for completeness and potential ethical concerns, as well as propose recommended classifications (i.e., 'Exempt', 'Expedited', or 'Full Board' categories). It could flag submissions that require further information (e.g., missing key information in the consent form) or that raise significant concerns (e.g., potential non-compliance with laws, regulations, or institutional policy; mention of vulnerable populations, and so on), allowing IRB staff to prioritize their review efforts. Similarly, the same function could also be extended to investigators to screen their draft applications, prior to IRB submission, to improve the quality and completeness of their submissions, while also flagging unnecessary jargon or potentially confusing passages of text.
b. **Preliminary review:** For protocols that pass the initial screening, the LLM/AI agent could autonomously generate a preliminary review report that: summarizes the study aims and methodology, key study design and ethical considerations; identifies relevant precedents and guidelines; and provides a tentative risk-benefit assessment. This report could serve as a *starting point* for the actual human review by the IRB member(s) or discussions at convened IRB meetings.
c. **Consistency checking:** The LLM, powered by a reasoning model, could be used to evaluate the IRB's proposed clarifications, modifications, or decisions with its past precedents and with relevant institutional policies and guidelines, and also recommendations and guidance from expert ethics/advisory committees. It could flag any potential inconsistencies or deviations, prompting the IRB to provide additional justification or to reconsider its approach. It could also flag inconsistencies among the protocol approved by grant funders, the actual protocol submitted for IRB approval, and any descriptions of the protocol contained in informed-consent information.
   d. **Decision support:** During the deliberation process, IRB members could interact with the LLM – or specifically agentic AIs, like 'Deep Research', that have access to both online and institutional resources – to ask for additional information, clarification, or analysis on specific issues. The LLM/agent could quickly retrieve relevant precedents, guidelines, and/or literature to generate a comprehensive fully cited report to inform the discussion.

## V. Discussion: Potential Benefits, Risks, and Replies

IRB-specific LLMs have the potential to address at least three of the most prominent concerns regarding the use of general LLMs in ethics review.

First, by being trained on and having access to additional information relevant to the specific tasks and context associated with an IRB's work, the accuracy of application-specific LLMs on review tasks is likely to be higher than that of more general models. By drawing on detailed knowledge bases established for the purpose, IRB-specific LLMs which 'reason' will be better able to provide relevant information and precedents, and less likely to generate inaccurate information.

Second, IRB-specific or expert LLMs may partially ameliorate concerns related to the lack of human empathy and judgment, since such LLMs are trained specifically on past examples of the use of human empathy and judgment in the context of research ethics evaluation.

Third, IRB-specific LLMs are trained on, and have access to, additional information concerning the national or local context of an IRB and its associated institution. This largely addresses concerns based on the need for IRB review to be sensitive to the national or local context in which the IRB operates. IRB-specific LLMs can make use of this information to adjust what would otherwise be context-independent recommendations. Significantly, if important information is missing, this can simply be fed into the vectorized knowledge base as needed.

While IRB-specific LLMs address several key limitations of the more general application of LLMs to IRB review, there are nevertheless multiple remaining issues that need to be acknowledged prior to any actual implementation of these ideas.

Even with careful curation and auditing, there is a risk that the IRB-specific datasets used to train the LLMs could reflect and perpetuate historical inequities or blind spots in the research enterprise. For example, if certain types of research or populations have been systematically underrepresented or excluded from past IRB reviews, the LLM may learn to discount or overlook their ethical significance. Similarly, if the training data contains implicit biases or stereotypes, the LLM may inadvertently reproduce these in its analyses and recommendations.

To mitigate these risks, it will be important to develop robust methods for assessing and mitigating bias in LLM training data and outputs. This could involve techniques such as testing, debiasing, and auditing, as well as ongoing monitoring and adjustment of the models in response to feedback from diverse stakeholders. [35] Post-deployment audits and reinforcement learning from human feedback, based on the differences between the LLM's reasonings/initial output and final IRB decisions integrating the human judgement, are key for continuous-improvement.

Another challenge is the potential for over-reliance on LLM outputs in the review process. While the goal is to use LLMs to support and enhance human decision-making, there is a risk that IRB staff and members could become overly deferential to the models' CoT reasonings, analyses, and recommendations, especially if they are presented as or otherwise seen as highly logical, accurate, or authoritative.[36] This could lead to a gradual erosion of the critical thinking and independent judgment that are essential to effective ethical oversight.

To counter this risk, it will be important to design the LLM-assisted review process in a way that actively engages and empowers human decision-makers. This could involve providing training and support to IRB staff and members on how to write effective prompts to instruct the LLM (i.e., user-level prompt engineering) and how to actively and critically interpret and respond to LLMs' CoTs and outputs, and including reminders in the LLM outputs to users that it is not a substitute for human judgement, and that the risk of hallucinations is always non-zero. It would also require establishing clear protocols for when and how human judgment should take precedence over machine recommendations.

Even with the accessibility of an LLM's CoT, there still remains important questions about the transparency and explainability of IRB-specific LLMs, and how these factors may impact the perceived legitimacy and trustworthiness of the review process. Given the high stakes and sensitive nature of research ethics, it is desirable that the reasoning behind LLM outputs can be clearly understood and interrogated by human reviewers and other stakeholders.[37,38] This issue can be partially mitigated through 1) application-specific IRB fine-tuning, 2) the extensive use of vectorized knowledge databases as it would then be clear what information was used to produce an output, and 3) system-level prompt engineering and design settings (e.g., low temperature; instruction to always find factual support based on the RAG knowledge base). Importantly, it could be primarily addressed through the use of systems specifically designed to be explainable, such as reasoners and their CoT. This in turn raises the question of how 'raw' or 'full' an LLM's thought processes should be to be considered sufficiently explainable, though this is a topic beyond the scope of this article.

Finally, we note that LLMs may have different utility in reviewing biomedical research protocols than in reviewing social/behavioral protocols. Risk in biomedical studies is often objective and quantifiable (e.g., the risk of a device-implantation surgery or the known side-effects of a study-drug). Social/behavioral risks are often more subjective, e.g., the risk that participation in a given study could harm a participant's reputation, or the level of privacy risk raised by participation in an online focus-group. Different IRBs can be expected to decide such questions differently, so the use of LLMs for study-screening and risk-assessment may not create uniformity across IRBs. On the other hand, the use of an LLM by a given IRB might decrease inconsistency of risk-judgments within that IRB over time, as the LLM gains "experience" through reinforcement learning through human feedback based on the actual decisions. LLMs could also be designed to suggest tools for risk-mitigation in social/behavioral studies, such as choosing meeting-places where the participant will not be seen meeting the researcher, or providing the researcher's contact information in a format that will not be recognized as research-related by participants' friends and family members, or using a particular approved mechanism for storing sensitive social data in the cloud.

## VI. Conclusion

Ultimately, the goal of this work is to foster a more ethical, efficient, and responsive system of research oversight that can keep pace with the rapidly evolving landscape of biomedical and social/behavioral research while upholding the highest standards of participant protection and scientific integrity. The background and reasons presented above convince us that IRB-specific LLMs at the very least show significant potential as tools towards the realization of this goal. Before anyone can know with certainty whether this is the case, much more theoretical and practical work remains to be done and pilot studies will need to be conducted and thoroughly evaluated. Nevertheless, the reasons laid out above convince us that this work represents a promising area worthy of these additional efforts.

## AI Use Acknowledgment

*Any use of generative AI in this manuscript adheres to ethical guidelines for use and acknowledgment of generative AI in academic research. Each author has made a substantial contribution to the work, which has been thoroughly vetted for accuracy, and assumes responsibility for the integrity of their contributions.*[39]